\documentclass[conference]{IEEEtran}
\IEEEoverridecommandlockouts

\usepackage{cite}
\usepackage{amsmath,amssymb,amsfonts}
\usepackage{algorithmic}
\usepackage{graphicx}
\usepackage{textcomp}
\usepackage{xcolor}
\usepackage{booktabs}
\def\BibTeX{{\rm B\kern-.05em{\sc i\kern-.025em b}\kern-.08em
    T\kern-.1667em\lower.7ex\hbox{E}\kern-.125emX}}
\begin{document}

\bstctlcite{IEEEexample:BSTcontrol}

\title{
Towards smart and adaptive agents for active sensing on edge devices \\
\thanks{\textsuperscript{*} Co-first authors. \textsuperscript{†} Co-senior authors. \\
\{devendra.vyas, miguel.deprado, tim.verbelen\}@verses.ai
}
\thanks{This work was partly supported by Horizon Europe dAIEdge under grant No. 101120726.}
}

\author{\IEEEauthorblockN{Devendra Vyas\textsuperscript{*}}
\IEEEauthorblockA{\textit{VERSES} \\
}
\and
\IEEEauthorblockN{Nikola Pižurica\textsuperscript{*}}
\IEEEauthorblockA{\textit{Computer Science Center, U. of Montenegro} \\
}
\and
\IEEEauthorblockN{Nikola Milović}
\IEEEauthorblockA{\textit{Fain Tech} \\
}
\and
\IEEEauthorblockN{Igor Jovančević}
\IEEEauthorblockA{\textit{Computer Science Center, University of Montenegro} \\
}
\and
\IEEEauthorblockN{Miguel de Prado\textsuperscript{†}}
\IEEEauthorblockA{\textit{VERSES} \\
}
\and
\IEEEauthorblockN{Tim Verbelen\textsuperscript{†}}
\IEEEauthorblockA{\textit{VERSES} \\
}
}

\maketitle

\begin{abstract}
TinyML has made deploying deep learning models on low-power edge devices feasible, creating new opportunities for real-time perception in constrained environments. However, the adaptability of such deep learning methods remains limited to data drift adaptation, lacking broader capabilities that account for the environment's underlying dynamics and inherent uncertainty. Deep learning's scaling laws, which counterbalance this limitation by massively up-scaling data and model size, cannot be applied when deploying on the \textit{Edge}, where deep learning limitations are further amplified as models are scaled down for deployment on resource-constrained devices.

This paper presents an innovative agentic system capable of performing on-device perception and planning, enabling active sensing on the edge. By incorporating active inference into our solution, our approach extends beyond deep learning capabilities, allowing the system to plan in dynamic environments while operating in real-time with a compact memory footprint of as little as 300 MB. We showcase our proposed system by creating and deploying a saccade agent connected to an IoT camera with pan and tilt capabilities on an NVIDIA Jetson embedded device. The saccade agent controls the camera's field of view following optimal policies derived from the active inference principles, simulating human-like saccadic motion for surveillance and robotics applications.  
\end{abstract}

\vspace{0.1cm}
\begin{IEEEkeywords} smart agents, edgeAI, dynamic planning

\end{IEEEkeywords}

\section{Introduction}
The human visual system has a unique ability to focus on key details within complex surroundings, a process known as saccading \cite{saccading}. This quick and dynamic scanning allows us to gather essential information. Saccading is part of a larger concept known as active (visual) sensing \cite{active_sensing, active_vision}, an innate capability that enables organisms to forage for information and dynamically adapt to an evolving environment~\cite{visual_foraging}.

\begin{figure}[t]
      \centering
      \includegraphics[width=0.95\linewidth]{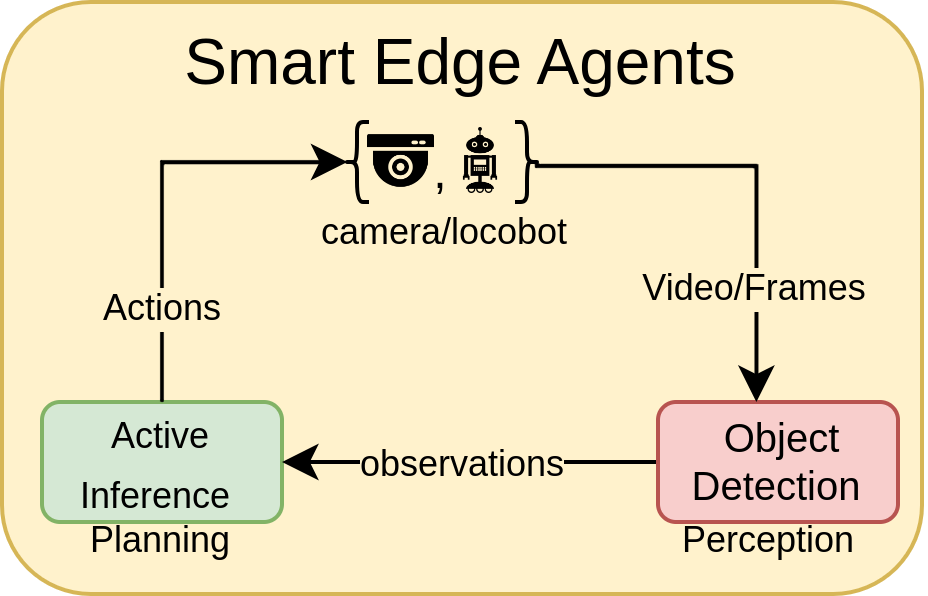}
      \caption{\textbf{Conceptual Framework for Smart Edge Agents}, composed of a deep-learning perception module and an active inference planning module for active (visual) sensing. The camera frames are processed by the object detector, which forwards the detected results to the active inference module. Our agent plans its next action, minimizing free energy, and dynamically adapting to the environment. }
      \label{fig:edge_saccading_agent}
  \end{figure}

Active sensing is critical in various applications, particularly when the information is unavailable or too vast to process. For instance, in remote sensing for Earth observation~\cite{earth_observation} or aerial search-and-rescue operations~\cite{search_rescue}, the system must parse vast, detailed scenes, focusing only on critical features, e.g., ice-sea or missing person. Similarly, in sports events, tracking dynamic scenes requires a system to zoom in on players, capturing players' faces or gestures while not missing the play. Other areas, such as smart cities and surveillance systems~\cite{smart_city}, demand robust monitoring solutions for crowded areas to track movement, anticipate potential issues, and enhance safety. Active sensing becomes even more apparent in robotics, where the agent's actions determine the next observations for the system, driving exploration~\cite{dacostaHowActiveInference2022a}.

Recent advances in machine learning (ML), particularly in deep learning, have substantially improved sensing accuracy and complexity. However, state-of-the-art deep learning models show limitations in their adaptability \cite{parisi2019continual}, i.e., the ongoing accumulation and refinement of knowledge over time. These limitations are further amplified when these models are scaled down for deployment on resource-constrained edge devices, where memory, computational power, and energy efficiency are limited. As a result, true active sensing—requiring both perception and planning—remains challenging to implement on embedded systems.

Active inference, an approach rooted in the first principles of physics, offers a promising alternative to address these limitations~\cite{ActInfBook}. Emerging as a viable paradigm, active inference grounds learning within probabilistic principles, enabling smart systems, or agents, to model the uncertainty and variability inherent in dynamic environments, making it well-suited for continual learning and adaptive decision-making~\cite{friston2016activelearning}. This shift represents a move beyond perception-focused AI toward adaptive systems capable of adjusting their actions based on environmental feedback.
Thus, agents on the edge provide a powerful framework for real-time perception and planning without dependence on cloud resources, ensuring low-latency responses and enhanced data privacy. 

This work presents an integrated system that combines deep learning and active inference to realize an adaptive, memory-efficient, real-time saccade agent for edge devices. Our system leverages a deep learning-based object detection module for initial perception and an active inference planning module to actively sense and adapt to the environment. The saccade agent can observe, plan, and control a camera for strategic information gathering, demonstrating adaptive decision-making and exploration. Our deployment on an Nvidia Jetson platform showcases the potential for responsive applications in robotics and smart city environments, highlighting the feasibility of edge-based adaptive systems for complex, real-world tasks.

\section{Related work}
We categorize the related work in three main areas:
\subsection{Active Sensing}
Active sensing is an essential building block across diverse fields where efficient scanning is needed to locate and focus on critical details. In Earth observation applications, the AutoICE Challenge \cite{earth_observation} addresses sea ice detection, where maximizing area coverage and detection through adaptive zooming is indispensable for safe navigation. Similarly, active search strategies are also explored in rescue operations~\cite{search_rescue}, stressing techniques like saccading to enhance search efficiency over large areas. For public safety in smart cities, deep learning methods are proposed for people tracking and counting~\cite{people_counting}, enabling security, crowd management, and urban analytics applications. 

\subsection{TinyML}
TinyML has made deploying ML models on low-power edge devices feasible, bringing opportunities for real-time perception in constrained embedded devices. The edge deployment pipeline is summarized in~\cite{prado2020bonseyes}, streamlining end-to-end model deployment on embedded platforms to enhance the accessibility of edgeAI applications. A popular example of this process is YOLO~\cite{yolo}, an efficient architecture for deploying object detection and tracking at the edge, improving the responsiveness of embedded applications. Recent works have made progress in enabling on-device domain adaptation, adjusting deployed applications to account for data distribution shifts between training and target environments~\cite {ElasticDNN,cioflan2024device}. However, these adaptations remain limited to addressing data drifts and lack broader capabilities for behavioral changes.
Our approach overcomes this limitation by integrating active inference on top of a deep learning module, allowing the system to plan and adapt to the environment accordingly. 

\subsection{Probabilistic computing}
Probabilistic computing has shown promise for active sensing by optimizing information acquisition in dynamic environments. Probabilistic principles can be used to maximize information gain through camera adjustment~\cite{sommerlade2008information}, which is valuable in applications like surveillance, sports analysis, and patient monitoring. This is extended to maximize mutual information gain in multi-camera setups, combining objectives like exploration and tracking to enable adaptive, informed scene monitoring~\cite{sommerlade2010probabilistic}. Unlike these methods, we base our probabilistic agent on active inference, grounding our approach in the Free Energy Principle.

\section{Methodology}
To develop an effective active sensing solution, it is essential to consider the unpredictable and dynamic nature of real-world environments. Smart sensors must be able to handle uncertainty and adapt to constant changes. Therefore, any change in the observed environment must influence the policy selection for the following action. For a system to operate autonomously and intelligently, it must be able to adjust its perception and actions in real time without relying on cloud processing. This requirement for on-device adaptation supports faster decision-making and enhances data privacy.

In this work, we propose an efficient active sensing agent composed of two modules: i) a deep learning-based perception module and ii) an active inference module that enables planning and control. This architecture combines deep learning’s feature extraction performance with active inference's Bayes-optimal control, presenting an adaptable and scalable solution for various resource-constrained edge applications.

\subsection{Perception}
Deep learning techniques have achieved remarkable success in detecting features of interest in images, audio, or textual data~\cite{lecun2015deep}. Convolutional neural networks (CNNs) have demonstrated exceptional performance in visual tasks like object detection and segmentation~\cite{object_detection_survey}. By employing deep learning for perception, active sensing agents can rapidly process high-dimensional data and identify patterns that provide relevant spatial information.

Recent advances in transformer architectures and large language models (LLMs) have further expanded the scope of deep learning. Transformers excel in capturing complex dependencies and long-range relationships in data, making them very powerful for tasks requiring a deeper contextual understanding. The self-attention mechanism, being a core component of transformers, enables the models to focus selectively on the most relevant aspects of the data, enhancing the agent's ability to perform active sensing by prioritizing critical visual cues.

However, while deep learning and LLMs offer powerful feature extraction, their static nature limits adaptability when deployed in uncertain or changing environments, which can only be counteracted by massively scaling data or model size. Thus, to address this challenge, our approach integrates deep learning for perception but relies on active inference as a much lighter, sample- and parameter-efficient higher-level module, enabling the agent to plan and dynamically adapt based on ongoing observations in real-time.

\subsection{Planning}
Active inference builds on the Free Energy principle, 
a theoretical framework stating that intelligent agents minimize the discrepancy between their internal (generative) model of the environment and incoming sensory data. By reducing this discrepancy, or "free energy", the agent maintains an updated and accurate representation of its surroundings, allowing the agent to make predictions about its environment and actively take actions to reduce uncertainty.

Concretely, the agent's generative model is a joint probability distribution over states $s$ and observations $o$. As inferring hidden states $s$ given some observations $o$ is typically intractable, an approximate posterior $Q(s | o)$ is introduced and optimized by minimizing the free energy $F$~\cite{ActInfBook}:

\begin{equation}
\min_{Q(s|o)} F = \underbrace{D_{KL}[Q(s|o)||P(s)]}_\text{complexity} - \underbrace{\mathbb{E}_{Q(s|o)}[\log P(o| s)]}_\text{accuracy}
\label{eq:F}
\end{equation}

Hence, the agent strives to give the most accurate predictions $P(o|s)$ while minimizing the complexity of the model with respect to the prior $P(s)$. To select actions or policies $\pi$ for the future $\tau$, an active inference agent will evaluate and minimize the expected free energy $G$ ~\cite{ActInfBook}:

\begin{equation}
\begin{aligned}
G(\pi) =& \underbrace{\mathbb{E}_{Q(o_\tau|\pi)}\big[D_{KL}[Q(s_\tau|o_\tau,\pi)||Q(s_\tau|\pi)]\big]}_{\text{(negative) information gain}}  \\ &- \underbrace{\mathbb{E}_{Q(o_\tau|\pi)}\big[\log P(o_\tau | C) \big]}_{\text{expected utility}}
\label{eq:G}
\end{aligned}
\end{equation}

The agent now averages across expected future outcomes $o_\tau$ and balances the expected information gain (i.e., exploration) with the expected utility (i.e., reward) encoded in prior preferences $C$. Both the observations $o$ and hidden states $s$ can be modeled as discrete variables with Categorical distributions, whereas optimizing $F$ and $G$ can be done using tractable update rules~\cite{friston2016activelearning}. Therefore, we convert the outputs of the deep learning perception module into a discrete observation space and use active inference for the action selection of our agent.



\section{Smart edge agent} \label{implementation}
Our smart edge agent, the saccade agent, comprises two modules, as introduced in the previous section, combining deep learning perception capabilities with active inference planning, as shown in Fig.~\ref{fig:edge_saccading_agent}. Specifically, we employ i) a deep-learning object detection model, offering efficient object and human detection capabilities directly at the edge, and ii) an active inference module that enables adaptive motion control (pan and tilt), allowing the agentic system to dynamically adjust its field of view or track detected entities autonomously. This adaptive behavior enhances the camera's utility as an intelligent surveillance IoT tool or for scene exploration and information foraging in robotic applications.
 
\subsection{Object detection using YOLOv10}
We chose YOLOv10~\cite{wang2024yolov10} from the YOLO family for our perception module due to its strong balance of detection accuracy and computational efficiency. YOLO models are single-pass object detectors that predict object categories and locations, making them ideal for real-time applications. YOLOv10
consistently demonstrates state-of-the-art performance and reduced latency 
across various model scales (N/S/M/L/X). We employ the Nano variant, with 2.3 million parameters, as it offers an efficient trade-off between accuracy, speed, and memory for efficient edge deployment.


To deploy the YOLOv10n network as efficiently as possible, we export it to ONNX~\cite{bai2019}. ONNX has become a standard for neural network representation and exchange and is widely supported by hardware vendors' software stacks, i.e., inference engines. This positions ONNX as a strong candidate for deployment space exploration and optimizations on a range of embedded devices. Thus, we create an edge-deployment workflow to find the most suitable deployment for YOLOv10n.

The edge-deployment workflow contains several edge-oriented inference engines, namely ONNX-runtime~\cite{onnxruntime}, TFlite~\cite{david2021tensorflow}, and TensorRT~\cite{jeong2022tensorrt} that can input an ONNX model and generate an optimized implementation for a given target hardware platform. These frameworks apply several optimizations across the network's graph, e.g., operator fusion, quantization, on the software stack, e.g., algorithm optimization, and leverage parallel hardware acceleration, vectorization, and optimized memory scheduling. The process results in a bespoke network description and a runtime that is ready for deployment on the hardware platform.

\subsection{Planning using Active Inference}
\begin{figure}[t!]
      \centering
      \includegraphics[width=0.8\linewidth]{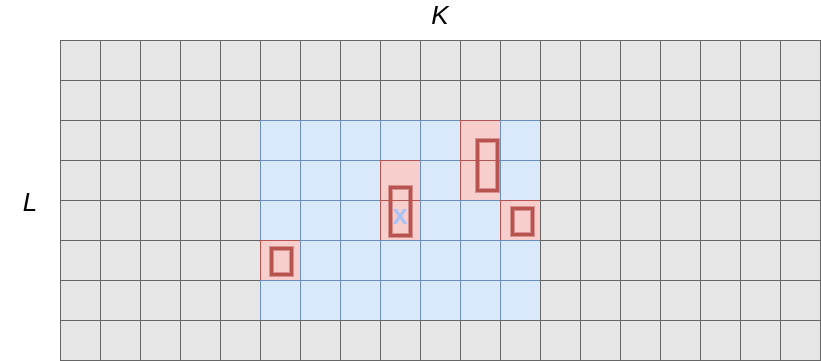}
      \caption{\textbf{Action space:} We discretize the action space into $K \times L$ fixation points. Given a fixation point, the field of view of the camera spans $W \times H$ blocks (in blue). Object detections are translated into discrete bins (in red).}
      \label{fig:saccade}
\end{figure}

To enable an efficient saccade agent with active inference planning, we define a discrete action, observation, and hidden state space. To this end, we divide the full area the camera can pan and tilt into a discrete grid of $K \times L$ blocks, see Fig.~\ref{fig:saccade}. Given a particular fixation point, the camera's field of view will only span $W \times H$ blocks, highlighted in blue. For each block, at each timestep, an observation $o_{w,h}$ is a Categorical distribution with three bins, i.e., the block can have no object detected (0 - blue), an object detected (1 - red) or not visible (2 - gray). The confidence of the bounding box outputs of the object detection neural network provides the probability of an object being detected. As state space, we similarly have a state variable $s_{k,l}$ per block, which is Bernoulli, i.e., object not present (0) or present (1). In addition, we also equip the agent with a proprioceptive state $s_p$ and observation, i.e., it observes the fixation point it is currently looking at. We specify the likelihood mapping \textbf{A} which predicts observation $o_{w,h}$ given state $s_{k,l}$ and fixation point $s_p$:

\begin{equation}
\textbf{A}_{w,h,k,l,p} = 
    \begin{cases}
    0 & \text{ if } s_{k,l} = 0, k\rightarrow_p w,  l\rightarrow_p h \\
    1 & \text{ if } s_{k,l} = 1, k\rightarrow_p w,  l\rightarrow_p h \\
    2 & \text{ otherwise}
    \end{cases}
\end{equation}

where $k\rightarrow_p w$ means ``block $k$ maps to observation $w$ given the agent looks at fixation point $p$''. We currently also assume that objects don't move considerably between timesteps and use the previous timestep posterior as the current prior, i.e., $P(s_t) = Q(s_{t-1}|o_{t-1})$, but we could expand this with dynamics modeling of the objects as future work.

The saccade agent uses the object detections received to perform inference, updating its beliefs about the hidden states. For example:
Detecting a ``person'' with high confidence would increase the probability assigned to the hidden state ``person present'' corresponding to the particular block. The absence of detection, on the other hand, would decrease the probability of that object being present.

After each observation, the agent evaluates possible actions, i.e., the next fixation points, based on their predicted outcomes. Actions are chosen to minimize expected free energy, which combines two key factors:
\begin{enumerate}
    \item \textbf{Expected observations matching prior preferences:} This essentially means choosing the most likely actions expected to lead to desired outcomes. For example, if the agent's goal is to locate a specific object, we set a preference $\textbf{C}_{w,h} = 1$, which favors actions that increase the probability of detecting that object in the field of view. Similarly, if we only set $\textbf{C}_{c_w,c_h} = 1$, with $(c_w, c_h)$ the center coordinates of the camera, this yields a ``tracking'' agent that pans/tilts to keep the object of interest in the center.
    \item \textbf{Epistemic value:} The agent also aims to reduce uncertainty about the hidden states. Actions expected to provide more informative observations about the environment have higher epistemic value (i.e., information gain in eq.~\ref{eq:G}). In our model, moving the field of view to previously unobserved blocks will result in a high amount of information gain. Hence, in the absence of objects of interest, the camera will pan and tilt to cover the whole area with as few moves as possible.
\end{enumerate}

We provide a qualitative demonstration of our saccade agent at \cite{demo}.

\section{Experimental results}

\subsection{Experimental Set up}
Our experimental setup comprises the following components, as shown in Fig.~\ref{fig:applications} :

\begin{figure}[t]
\centering
{\includegraphics[width=0.2\textwidth] {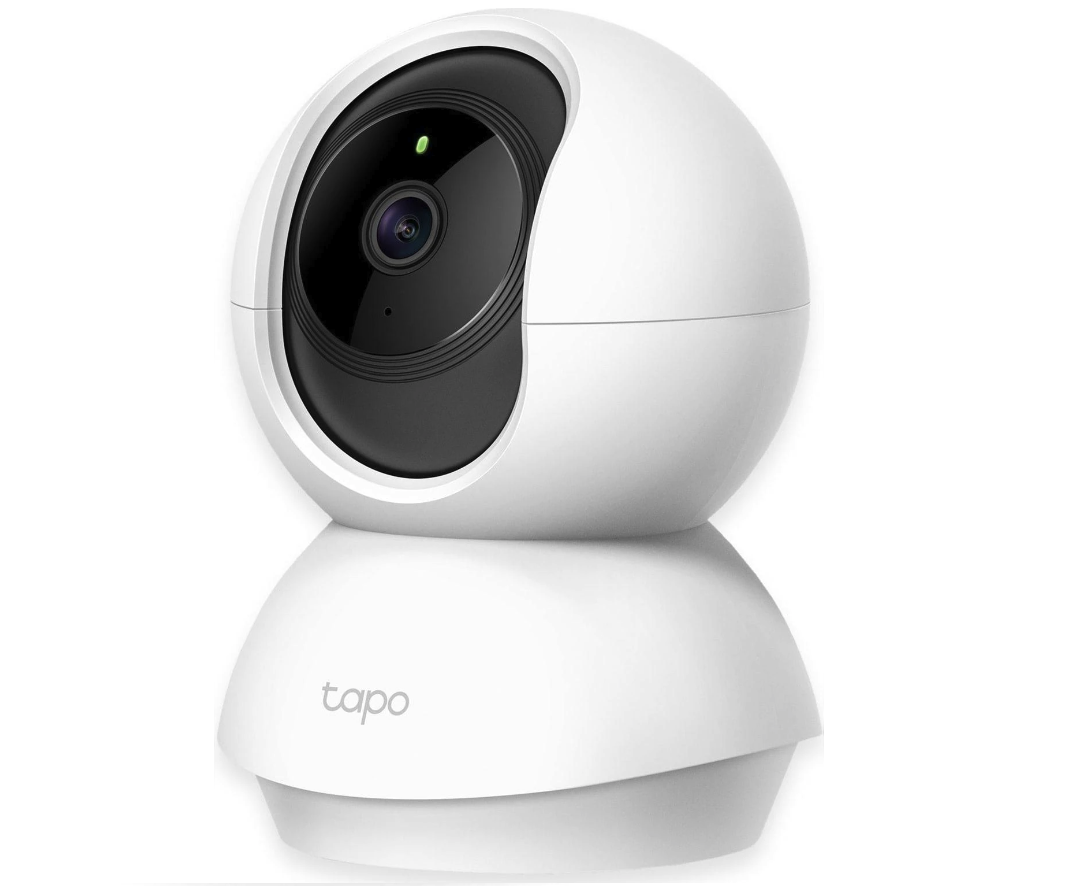}\label{fig:camerito}}
\hfill
{\includegraphics[width=0.25\textwidth]{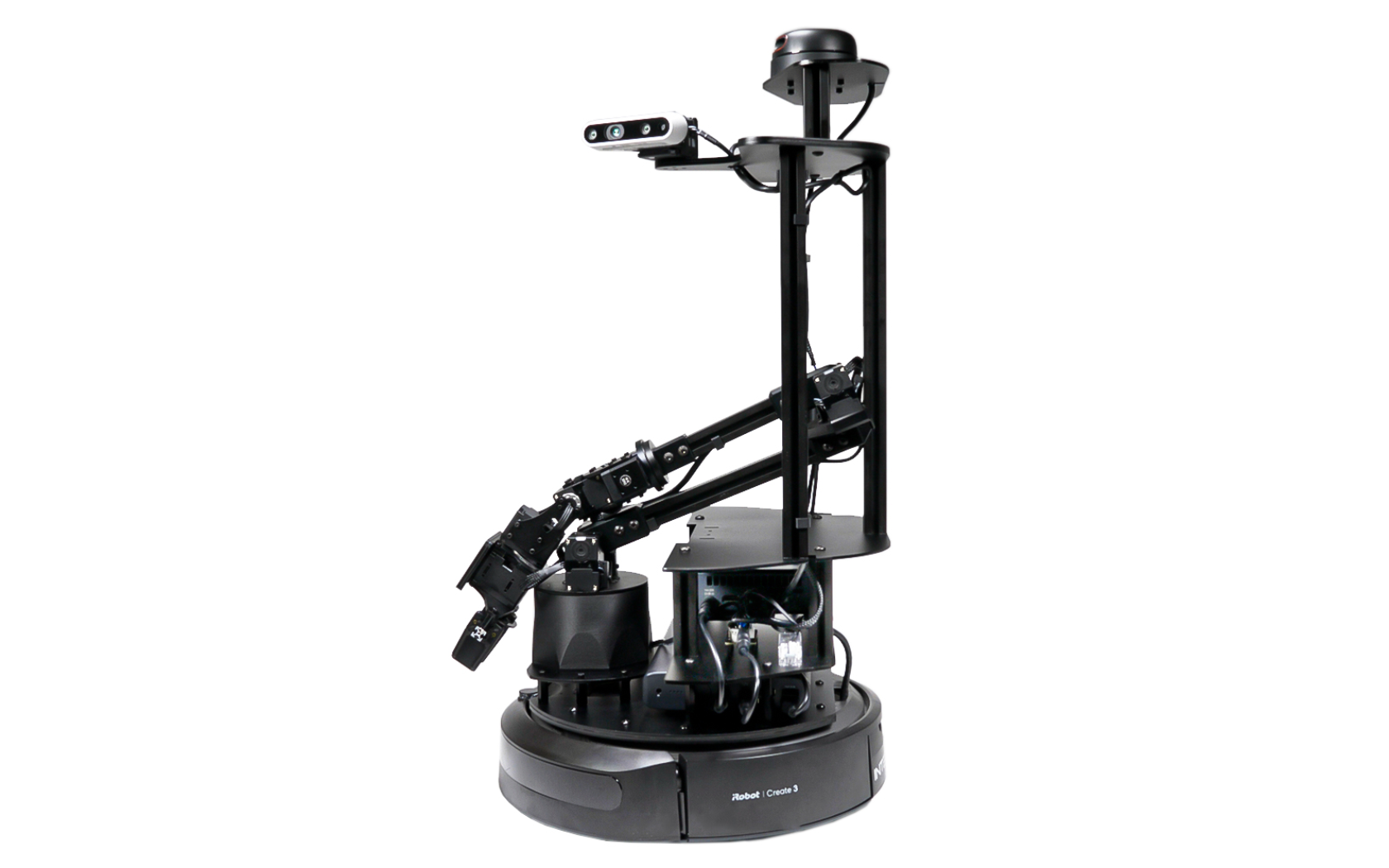}\label{fig:karlito}}
\caption{\textbf{Applications:} Our agentic system enables active sensing solutions for edge robotics and surveillance IoT cameras. On the left, the Tapo IoT camera~\cite{tapo} used for surveillance applications. On the right, the Locobot robot WX250~\cite{locobot} for information gathering and scene discovery.}
\label{fig:applications}
\end{figure}

\begin{enumerate}
    \item \textbf{Tapo Camera:} We use an IoT Tapo camera equipped with pan and tilt capabilities, which serves both as the input for the observations, i.e., images/frames, forwarded to the object detector, and the actuator for our saccade agent. The agent commands the camera to perform dynamic adjustments in its field of view based on optimal policies derived by minimizing free energy, simulating a human-like saccadic motion to maintain focus on areas of interest. 
    \item \textbf{Locobot WX250:} We also deploy our agent on the Locobot WX250, a mobile robot platform for autonomous mobility. Equipped with a 6-DOF manipulator and pan/tilt camera, the Locobot enables active exploration of the environment. This capability complements the Tapo camera's pan and tilt actions with navigation. The robot's mobility enhances the agent's capacity for active sensing in new environments, allowing it to reposition itself and collect additional perspectives to refine perception in complex settings.  
    \item \textbf{Nvidia Jetson Orin NX:} Our setup leverages the Nvidia Jetson Orin NX. Equipped with an 8-core ARM Cortex-A78 CPU and a 1024-core Ampere GPU, the Jetson Orin NX represents a powerful edge AI platform designed for accelerated machine learning tasks with up to 100 TOPs of processing capability and 16 GB of memory, running at 25W. 
\end{enumerate}

A pre-trained YOLOv10n model on the COCO dataset~\cite{lin2014microsoft} is deployed on the Nvidia Jetson Orin NX for real-time detection. The active inference module receives bounding boxes as observations and returns the optimal pan and tilt actions for the IoT or robotic camera. Both perception and planning modules employ the edge-deployment workflow to find the most suitable deployment engine, yielding the highest performance. 

\subsection{Results}
Next, we summarize our deployment experimental results: 
\subsubsection{Perception} 
\begin{figure}[t!]
      \centering
      \includegraphics[width=0.95\linewidth, height=6cm]{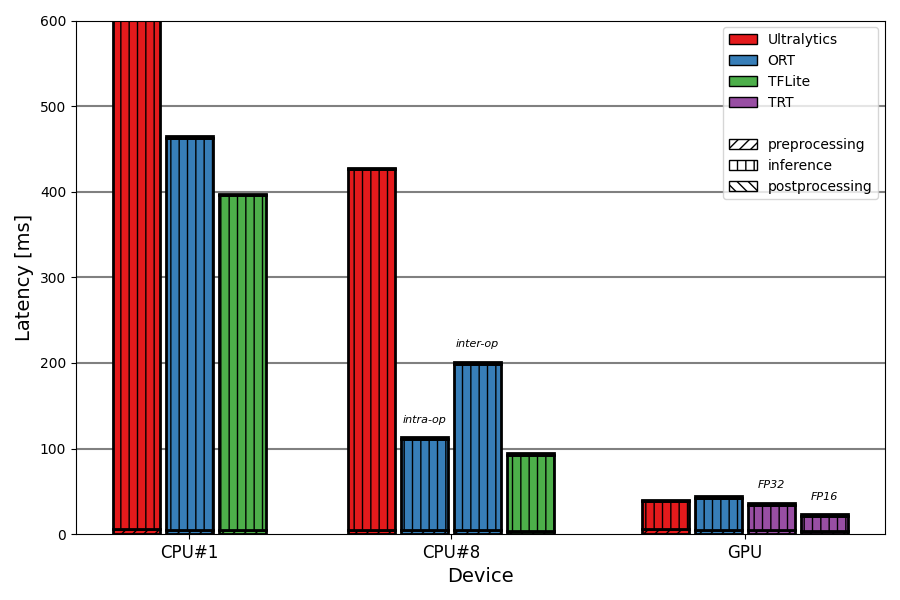}
      \caption{\textbf{Perception module performance:} Optimization achieved by exporting the YOLOv10n Torch model from Ultralytics to ONNX and compiling it with different deployment inference engines on the Nvidia Jetson Orin NX's CPU and GPU.\vspace{-0.2cm}}
      \label{fig:perception-latency}
\end{figure}

We optimized the YOLOv10n model, as introduced in Section ~\ref{implementation}, by exporting the original Torch model from the original \textit{Ultralitycs} library to ONNX. Then, the model gets compiled with one of the various deployment inference engines. Figure~\ref{fig:perception-latency} depicts the average results of 1000-inference runs, with one warm-up sample, while deploying the different implementations on the Nvidia Jetson's CPU, single- and multi-core, and GPU. 

When evaluating the various inference engines using single-floating-point precision (FP32) operations on a single-core CPU, Ultralitics comes out as the slowest inference, with over 600 ms. When the model is exported to ONNX and compiled with ONNX-runtime (ORT) and TensorFlow Lite (TFLite), the latency decreases considerably by 23\% 35\%, respectively. We obtain higher gains when parallelizing across all available 8 CPU cores, achieving 427 ms when using Ultralytics. The performance can be notably increased if we deploy with ORT (intra-operator parallelization) and TFLite, with a 3.75x and 4.5x speed-up over Ultralytics, respectively. Finally, when offloading inference to the GPU, we see a boost in performance for all GPU-available frameworks by leveraging the native parallel compute units of the device, achieving an inference in under 45 ms. When using TensorRT (TRT), designed explicitly for Nvidia Jetson devices, we accomplish an optimized deployment through reduced numerical precision (FP16), with as little as 22 ms, i.e., 45 FPS. 

\begin{figure}[t!]
      \centering
      \includegraphics[width=0.95\linewidth, height=6cm]{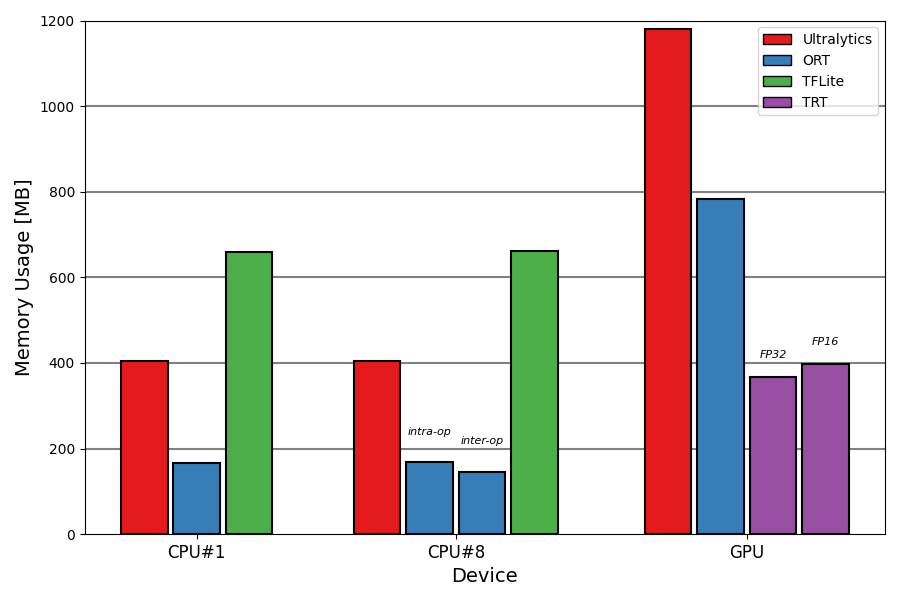}
      \caption{\textbf{Perception module memory profile} of the YoloV10n when executed with different deployment inference engines on the Nvidia Jetson Orin NX's CPU and GPU.\vspace{-0.2cm}}
      \label{fig:perception-memory}
\end{figure}

Figure~\ref {fig:perception-memory}  shows the memory profile of the YOLOv10n deployments with the various inference engines. It can be observed that while TFLite provides the fastest deployment on the single-CPU setup, it also consumes the most memory, with over 50\% more than Ultralytics. On the other hand, ORT provides a good balance between latency and memory utilization, being slightly slower than TFLite, but achieving inference with as little as 168 MB. This pattern is also repeated in the multi-CPU deployment setup. On the GPU side. Ultralytics, being a training-oriented framework, is the one that consumes the most memory, with up to 1.2 GB. As we discussed earlier, TRT is designed explicitly for Nvidia Jetson devices. As such, it accomplishes a very performant deployment with an optimized memory usage under 400 MB.

Overall, the results demonstrate that the predominant contribution to overall latency stems from the neural network inference stage. In contrast, the pre-processing phase exhibits minimal computational overhead, and the post-processing step remains negligible — a characteristic intrinsic to the YOLOv10 architecture. Moreover, the comparative analysis across inference engines and hardware configurations reveals distinct performance and memory usage behaviors. For instance, TFLite achieves favorable latency on single-core execution, although with the highest memory usage, while ORT offers an excellent performance-memory tradeoff. On the GPU side, TRT is the absolute winner with excellent latency-memory performance. These observations highlight the importance of a flexible edge-deployment workflow that systematically identifies the most suitable software–hardware pairing for a given operational context. 

\subsubsection{Planning}
Next, we deploy the active inference planning module, which only contains 1.34K parameters, on the Nvidia Jetson NX. The original active inference model was designed with the \textit{Pymdp} library~\cite{heins2022pymdp}, which contains a JAX backend. We then follow a similar workflow as with YOLOv10n and export the active inference model to ONNX and deploy it with several inference engines.

\begin{figure}[t!]
      \centering
      \includegraphics[width=0.95\linewidth]{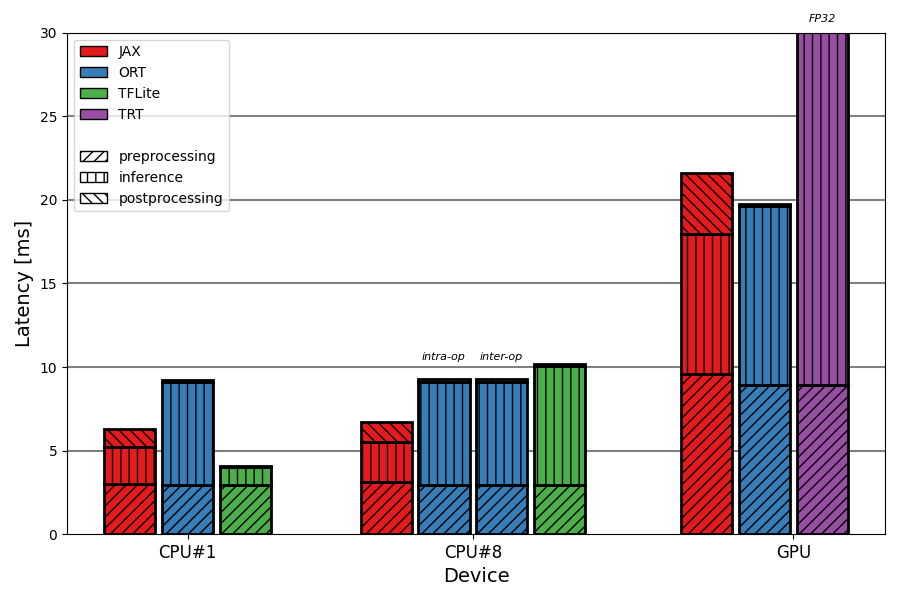}
      \caption{\textbf{Planning module performance}: Performance optimization by exporting the active inference model (saccade agent) from JAX to ONNX and compiling it with different deployment inference engines on the Nvidia Jetson Orin NX’s CPU and GPU.}
      \label{fig:saccade-latency}
\end{figure}

Figure~\ref{fig:saccade-latency} depicts the latency of the active inference model when planning the following position to look at. The pre-processing step accounts for decoding and mapping the object detection bounding boxes to the generative model. The inference part involves the \textit{infer\_states} method, which updates the agent's beliefs about the hidden states given the observations, and the \textit{infer\_policies} method, which defines the set of policies from which the agent will pick the best action. When deployed on a single CPU, the planning process executes fast, achieving 6 ms using JAX and 4 ms (250 FPS) when compiled with TFLite. The saccade agent does not benefit from parallel processing, showing similar or worse performance when deployed on multiple CPUs and considerably worse when offloaded to the GPU. We argue this is due to the small size of the agent, paying a high penalty for thread synchronization, which dominates over the benefit from parallel computation.

\begin{figure}[t!]
      \centering
      \includegraphics[width=0.95\linewidth]{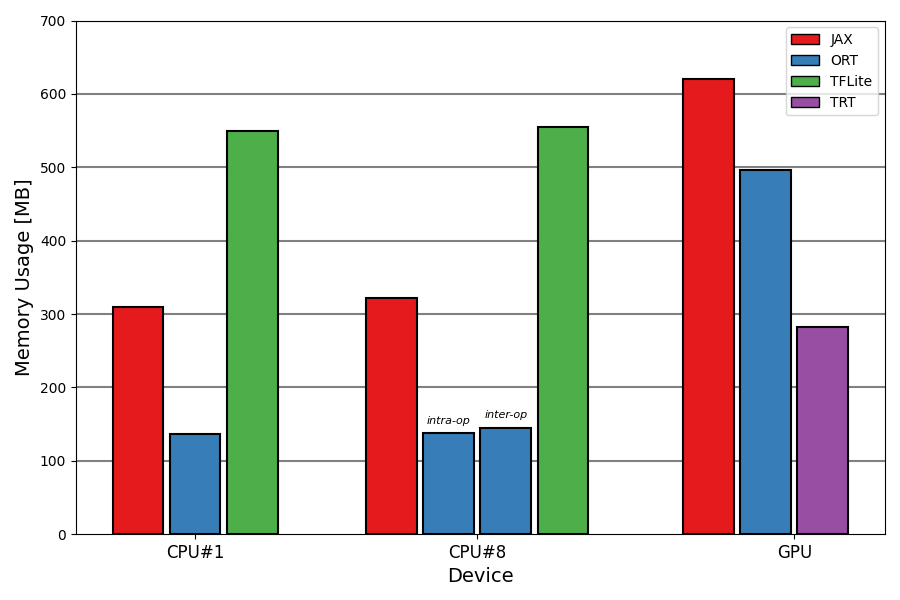}
      \caption{\textbf{Planning module memory profile} of the active inference model when executed with different deployment inference engines on the Nvidia Jetson Orin NX CPU and GPU. \vspace{0.3cm}}
      \label{fig:saccade-memory}
\end{figure}

Figure~\ref{fig:saccade-memory} illustrates a trend consistent with that observed for the YOLOv10n model, revealing a trade-off between inference performance and memory consumption. For example, while TFLite delivers the fastest execution, it may consume up to four times more memory than ORT. This behavior may be related to internal optimization mechanisms in TFLite, potentially involving intermediate data caching or buffer reuse strategies to reduce computation time.

Overall, these results highlight the lightweight deployment of our active inference model, achieving real-time planning and adaptation to the environment for IoT and robotics saccading applications.

\subsubsection{System} Integrating the perception and planning modules produces a highly efficient saccade agent capable of real-time operation with only 2.3 million parameters. Table 1.1 summarizes key configurations, illustrating the trade-offs between latency and memory consumption. For example, in configuration \textit{A}, the perception module optimized with TensorRT-FP16 achieves a throughput of 45 FPS, enabling rapid feature extraction. In parallel, the active inference module using TFLite can perform planning and decision-making at up to 250 FPS, with a total memory footprint under 1 GB. The decision-making rate can be reduced to 108 FPS using ORT, lowering memory usage to 533 MB, as shown in configuration \textit{B}. Finally, configuration \textit{C} employs the ORT (intra-operator parallelization) for perception, decreasing the perception rate to 9 FPS while providing a compact memory footprint of 304 MB, including all required libraries and runtimes. This configuration effectively balances perception and decision-making’s latency and memory, enabling dynamic adaptation and efficient operation on resource-constrained edge devices.

\begin{table}[t!]
    \centering
    \begin{tabular}{
     p{0.20\linewidth}
     p{0.20\linewidth} 
    p{0.20\linewidth} 
    p{0.25\linewidth}} 
        \centering\# Params (M) & \centering Memory (GB) &  \centering Perception Rate (Hz) &  Decision-making Rate (Hz)  \\
        \toprule
         A\hspace{0.5cm}2.3 & \hspace{0.5cm}947 & \hspace{0.7cm}45 & \hspace{0.5cm}\textbf{250}  \\
         B\hspace{0.5cm}2.3 & \hspace{0.5cm}533 & \hspace{0.7cm}\textbf{45} & \hspace{0.5cm}108 \\
         C\hspace{0.5cm}2.3 & \hspace{0.5cm}\textbf{304} & \hspace{0.85cm}9 & \hspace{0.5cm}108 \\ \vspace{0.2cm}
    \end{tabular}
    \caption{\textbf{System parameters}: Saccade agent deployed on the Nvidia Jetson Orin NX\vspace{-0.7cm}}
    \label{tab:system_table}
\end{table}

\subsection{Discussion}
These results highlight the effectiveness of combining deep learning for perception with active inference for planning in edge-based intelligent agents. With only 1.34K parameters, the active inference module achieves real-time planning while maintaining minimal computational and memory demands. In contrast to large-scale foundation models that rely on billions of parameters to achieve general-purpose intelligence, our domain-oriented approach demonstrates how compact, task-specific active inference models can complement deep learning perception. The adaptability of our edge-deployment workflow illustrates deployment interoperability, facilitating the exploration of diverse configurations to achieve a suitable balance between latency, accuracy, and resource efficiency.


\section{Conclusions and Future Work}
This work has introduced a smart edge agent composed of a deep learning-based perception module and an active inference planning module for active sensing. The system demonstrates the feasibility of adaptable, on-device surveillance and robotic solutions and their potential to handle real-world challenges effectively. Our study highlights the potential of active inference in edge-based systems. Active inference offers a robust framework that accounts for the environment’s underlying dynamics and inherent uncertainty. It allows the agent to actively reduce ambiguity or explore the environment for new information. This approach establishes a foundation for efficient edge agents for adaptive, resource-efficient decision-making in IoT and edge computing applications.

In future work, we aim to compare the proposed system against other state-of-the-art works, showing a quantitative functional comparison on popular challenges, such as the Habitat Navigation Challenge. Finally, we envision extending this concept to other applications, which could unlock a new generation of adaptive edge devices capable of context-driven interactions in complex environments.


\bibliographystyle{IEEEtran}

\bibliography{bibs}

\end{document}